%% file: main.tex
\documentclass[lettersize,journal]{IEEEtran}
\usepackage{amsmath,amsfonts}
\usepackage{algorithmic}
\usepackage{algorithm}
\usepackage{array}
\usepackage[caption=false,font=normalsize,labelfont=sf,textfont=sf]{subfig}
\usepackage{textcomp}
\usepackage{stfloats}
\usepackage{url}
\usepackage{verbatim}
\usepackage{graphicx}
\usepackage{cite}

\usepackage{multirow}
\usepackage{graphics}
\usepackage{epsfig}
\usepackage{mathptmx} 
\usepackage{times}
\usepackage{amssymb}
\usepackage{mathrsfs}
\usepackage{color}
\usepackage{amsthm}
\usepackage[cal=cm]{mathalfa}
\usepackage[pagebackref,breaklinks,colorlinks]{hyperref}

\begin{document}

\title{RefHCM: A Unified Model for Referring Perceptions in Human-Centric Scenarios}

\author{Jie~Huang,
        Ruibing~Hou$^\dag$,~\IEEEmembership{Member,~IEEE},
        Jiahe~Zhao,
        Hong~Chang,~\IEEEmembership{Member,~IEEE}, 
        Shiguang~Shan,~\IEEEmembership{Fellow,~IEEE}
\thanks{\dag~Corresponding author.}
\thanks{This work is supported in part by Natural Science Foundation of China (NSFC) under Grants 62306301, and in part by National Postdoctoral Program for Innovative Talents under Grant BX20220310.}
\thanks{Jie Huang, Jiahe Zhao, Hong Chang and Shiguang Shan are with Key Laboratory of Intelligent Information Processing, Institute of Computing Technology (ICT), Chinese Academy of Sciences (CAS), Beijing, 100190, China, and University of Chinese Academy of Sciences, Beijing, 100049, China. (e-mail: \{huangjie24s, zhaojiahe22s, changhong, sgshan\}@ict.ac.cn)}
\thanks{Ruibing Hou is with Key Laboratory of Intelligent Information Processing, Institute of Computing Technology (ICT), Chinese Academy of Sciences (CAS), Beijing, 100190, China. (e-mail: houruibing@ict.ac.cn)}
}

\markboth{Journal of \LaTeX\ Class Files,~Vol.~14, No.~8, August~2021}%
{Shell \MakeLowercase{\textit{et al.}}: A Sample Article Using IEEEtran.cls for IEEE Journals}

\IEEEpubid{0000--0000/00\$00.00~\copyright~2021 IEEE}

\maketitle

\input{0_abstract}

\input{1_introduction}

\input{2_related_work}

\input{3_method}

\input{4_experiments}

\input{5_conclusion}

\bibliographystyle{IEEEtran}
\bibliography{7_refer}

\end{document}

%% file: 0_abstract.tex
Human-centric perceptions play a crucial role in real-world applications. While recent human-centric works have achieved impressive progress, these efforts are often constrained  to the visual domain and lack interaction with human instructions, limiting their applicability in broader scenarios such as chatbots and sports analysis. This paper introduces \textit{Referring Human Perceptions}, where a referring prompt specifies the person of interest in an image. 
To tackle the new task, we propose RefHCM (\textbf{Ref}erring \textbf{H}uman-\textbf{C}entric \textbf{M}odel), a unified framework to integrate a wide range of  human-centric referring tasks. Specifically, RefHCM employs sequence mergers to convert raw multimodal data—including images, text, coordinates, and parsing maps—into semantic tokens. This standardized  representation enables RefHCM to reformulate diverse  human-centric referring tasks into a sequence-to-sequence paradigm, solved using a plain encoder-decoder transformer architecture. Benefiting from a unified learning strategy, RefHCM effectively facilitates knowledge transfer across tasks and exhibits  unforeseen capabilities in handling complex reasoning.
This work represents the first attempt to address referring human perceptions with a general-purpose framework, while simultaneously establishing a corresponding benchmark that sets new standards for the field. 
Extensive experiments showcase RefHCM's competitive and even superior performance across multiple  human-centric referring tasks.
The code and data are publicly at \href{https://github.com/JJJYmmm/RefHCM}{https://github.com/JJJYmmm/RefHCM}.

\vspace{2mm}


%% file: 1_introduction.tex
\section{Introduction}

Human-centric perceptions play an important role in widespread applications, including  augmented reality \cite{zhenwang2021ar, kaiyuanhu2023ar}, sports analytics \cite{tom2019sport, yutaro2022sport} and AI-generated content \cite{yao2023video, lihu2023video}. 
This field spans massive critical tasks such as pose estimation \cite{gregory2020pose, taoyu2020pose, yiyang2013pose, haoshufang2023pose}, pedestrian detection, pedestrian attribute analysis \cite{jianjia2021attribute, jianjia2022attribute, wanhuali2022attribute, haoxuanyou2024attribute}, and human parsing \cite{sanyizhang2022parsing, ziweizhang2022parsing, xiaodanliang2016parsing}. However, most existing human-centric models are specialized for individual tasks, resulting in significant costs associated with network design and parameter tuning. To enable efficient and scalable real-world deployment, it is important to develop a foundation  model that can be adapted to various human-centric perceptions tasks.

There are two mainstream approaches to developing human-centric foundation models. One line follows 
pretraining to fine-tuning paradigm. In this line, human representations are typically pretrained through  self-supervised techniques \cite{weihuachen2023solider,wentaozhu2023motionbert} or multi-task supervised pretraining. However, this paradigm still requires fine-tuning for each specific downstream task, resulting in heavy workload for fine-tuning and model development.
Another line focuses on developing a multi-task human-centric model. These efforts  \cite{yuanzhengci2023unihcp,yizhouwang2023hulk} aim to unify multiple human-centric tasks within a single model. While promising, these approaches design distinct loss functions for different tasks, which can lead to conflicts and complicate the balancing of tasks. 
Additionally, these methods struggle with handling multimodal inputs and outputs of flexible lengths, limiting their ability to interact naturally with humans. 

Different from previous works \cite{yuanzhengci2023unihcp,weihuachen2023solider}, our work focuses on developing a unified \emph{referring} human-centric model that predicts human  perceptions of a referred individual using user-friendly text.
Specifically, we  begin by focusing on two extensively studied referring tasks: Referring Expression Comprehension (\emph{REC}) \cite{lichengyu2016rec} and Referring Expression Generation (\emph{REG})\cite{lichengyu2016rec}. REC involves locating visual regions based on textual descriptions, while REG generates brief linguistic expressions for specified visual regions.
Building upon these, we introduce three referring tasks tailored to human-centric scenarios: Referring Keypoint (\emph{RKpt}), Referring Parsing (\emph{RPar}) and Referring Human-Related Caption (\emph{RHrc}).
\emph{RKpt} and \emph{RPar} aim to generate keypoints and parsing maps, respectively, for individuals specified by input  descriptions. \emph{RHrc} emphasizes generating detailed human-centric captions that go beyond brief expressions in \emph{REG}, offering richer and more comprehensive descriptions of the specified visual regions.
We aim to explore the underlying homogeneity across these tasks to build a human-centric  foundation model  capable of seamlessly address a wide range of   referring tasks.

\IEEEpubidadjcol

Two main challenges need to be solved for building such a foundation model. The first challenge lies in creating a unified representation space across different human-relevant modalities.  The inputs and outputs of various referring tasks span heterogeneous  formats, \textit{e.g.}, image, languages, bounding box, keypoints and parsing map. 
Unifying these highly diverse inputs and outputs  into a single cohesive representation space is nontrivial for a foundational model. 
The second challenge involves designing a unified network architecture and optimization objective that can seamlessly handle different human-centric referring tasks. Existing approaches use different network architectures and objective functions for various tasks. For instance, HRFormer \cite{yuhuiyuan2021hrformer} is designed for pose estimation, and CE2P \cite{taoruan2019ce2p} is used for parsing. Different tasks also employ distinct loss functions, such as regression loss for parsing and classification loss for text generation.
Designing task-specific architectures and objectives is often labor-intensive and may cause conflicts between tasks.
Therefore, unifying these specialized architectures and objectives is crucial to building an effective referring foundation model.

In this paper, we propose a unified \underline{Ref}erring \underline{H}uman-\underline{C}entric \underline{M}odel, namely RefHCM, that unifies various referring human-centric tasks into a sequence-to-sequence paradigm.
RefHCM comprises three main tries.
\textbf{Firstly}, to establish a unified representation space across diverse human-relevant modalities, RefHCM is equipped with sequence mergers and dispensers. The mergers are capable of merging raw multimodal data, including image, text, spatial coordinates, and parsing map, into a cohesive sequence of \textit{tokens}. The dispensers then separate the output sequence into modality-specific segments and convert each segment into its respective output format.
\IEEEpubidadjcol
\textbf{Secondly}, RefHCM reformulates diverse human-centric referring tasks into a unified \textit{sequence-to-sequence paradigm} by leveraging the mergers and dispensers.
Specifically,
a modality-agnostic encoder integrates tokens from input modalities, including images and user instructions.
These integrated vision-language tokens are subsequently processed by a causal decoder, which  generates responses autoregressively based on the aggregated tokens.
\textbf{Lastly}, RefHCM adapts to the specific characteristics of different human-centric tasks.
For \emph{REC} and \emph{RKpt}, we introduce  a \textit{Location-Context Restriction} mechanisms that leverages the mutual optimization between human bounding box and keypoint information, enhancing both location and pose estimation.  
For \emph{RPar}, we propose \textit{Query Parallel Generation} (QPG), which combines autoregressive and parallel generation manners to address latency issue in purely  autoregressive generation. 
QPG achieves a 48x acceleration in parsing map generation while preserving 88\% of the original performance.

To further evaluate the model's reference reasoning capabilities, we introduce a  \underline{Reason}ing \underline{Ref}erence  (\emph{ReasonRef}) Benchmark. This benchmark challenges the model to predict human perceptions—including bounding boxes, keypoints, and parsing maps—based on implicit references that require complex reasoning. 
Unlike straightforward and direct references, the referring texts in \emph{ReasonRef} incorporate more  intricate descriptions that demand advanced interpretative and reasoning skills. 
The reasoning tasks are categorized into five dimensions: \textit{Identity}, \textit{Pose/Clothing}, \textit{Social Relations}, \textit{Physical Relations} and \textit{Future Prediction}. This comprehensive categorization provides a
detailed assessment of the model's ability to understand and reason  across diverse aspects of human-centric perceptions.

We evaluate RefHCM on the coco dataset family, CIHP and \emph{ReasonRef} benchmarks to assess its capability  in understanding both simple references and complex reasoning tasks. Extensive experiments demonstrate that RefHCM achieves competitive performance across multiple human-centric referring tasks. Additionally, we demonstrate that RefHCM exhibits impressive zero-shot generalization to complex reasoning tasks, despite being trained solely on simple, direct references.

%% file: 2_related_work.tex
\section{Related Work}

\subsection{Human-centric Perception Models} 
Human-centric perception models play a crucial role in real-world applications, such as social surveillance and sports analysis. This field can  be broadly categorized into three mainstream approaches:

\vspace{2mm}
\noindent
\textbf{Task-Specific Models.} \
Task-specific models are designed to tackle individual tasks, focusing on improving network architecture and loss functions for a specific task. 
For instance, in pose estimation, HRNet \cite{bowencheng2020hrnet} fuses multi-resolution information to precisely  detect  fine-grained human keypoints, while DEKR \cite{ziganggeng2021dekr} directly regresses keypoints, bypassing  the need for heatmap \cite{junjiehuang2020heatmap} optimization. 
In human parsing, several models \cite{kegong2019gcn} employ  Graph Convolutional Networks to enhance information exchange between body parts. 
While  these task-specific models often achieve state-of-the-art results, deploying multiple models for multi-task applications presents  significant challenges in terms of efficiency and scalability.

\vspace{2mm}
\noindent
\textbf{Human-centric Pretraining Models.} \
This line incorporates human priors to pretrain a versatile backbone for extracting general human-centric representation. Specifically,  
SOLIDER \cite{weihuachen2023solider} enriches human representation by embedding additional semantic information through a  semantic classification pretext task. 
HAP \cite{junkunyuan2023hap} utilizes offline-extracted human keypoints as prior knowledge to learn structure-invariant representation across various human poses. 
Despite these advancements, these pretraining models still require  finetuning for each specific downstream task, limiting their flexibility.

\vspace{2mm}
\noindent
\textbf{Unified Human-centric Models.} \
Recently, several approaches have aimed at developing a unified framework to handle diverse human-centric tasks. For instance, 
UniHCP \cite{yuanzhengci2023unihcp} adopts a DETR\cite{nicolas2020detr} architecture, which introduces a shared decoder head alongside task-specific interpreters to handle five human-centric vision tasks.   
HULK \cite{yizhouwang2023hulk} employs a similar structure, extending support to additional tasks such as 3D pose estimation and short caption generation.
UniPHD \cite{uniphd2024miao} improves deformable DETR \cite{deformable_detr2021zhu} for pose processing, enabling simultaneous prediction of human poses and instance masks from either natural language descriptions or position-based prompts.
These DETR-like structures enable parameter sharing across tasks, exploiting inter-task homogeneity to enhance overall performance. However, most above approaches lack referring understanding capabilities, restricting their applicability in human-computer interaction scenarios such as augmented reality.

\subsection{Encoder-Decoder Language Models}

The encoder-decoder architecutre is exemplified by models such as T5 \cite{colin2020t5} and BART \cite{mike2020bart}.
Unlike prevalent decoder-only language models, T5 divides text into two parts—typically instruction and response—which are then fed into the encoder and decoder, respectively. 
Models like Florence-2 \cite{binxiao2023florence} and OFA \cite{pengwang2022ofa} extend this framework to multimodal scenarios by incorporating additional visual encoders and resamplers, aligning visual and textual information as part of the instruction to the encoder.
We further adapt this encoder-decoder architecture for human-centric tasks with several key modifications. Specifically, we integrate a modified VQGAN \cite{patrick2021vqgan} decoder, which enables the generation of multichannel dense parsing maps.  
 Additionally, we introduce a query-autogresssive generation strategy that  combines query-based and autoregressive methods to address the latency challenge associated with purely  autoregressive generation. 
  These modifications enhance the model's ability to handle diverse referring perception tasks within a unified framework.

%% file: 3_method.tex
\section{Referring Human-centric Tasks}
\label{sec3}
\begin{figure*}[htbp]
    \centering
    \includegraphics[width=\textwidth]{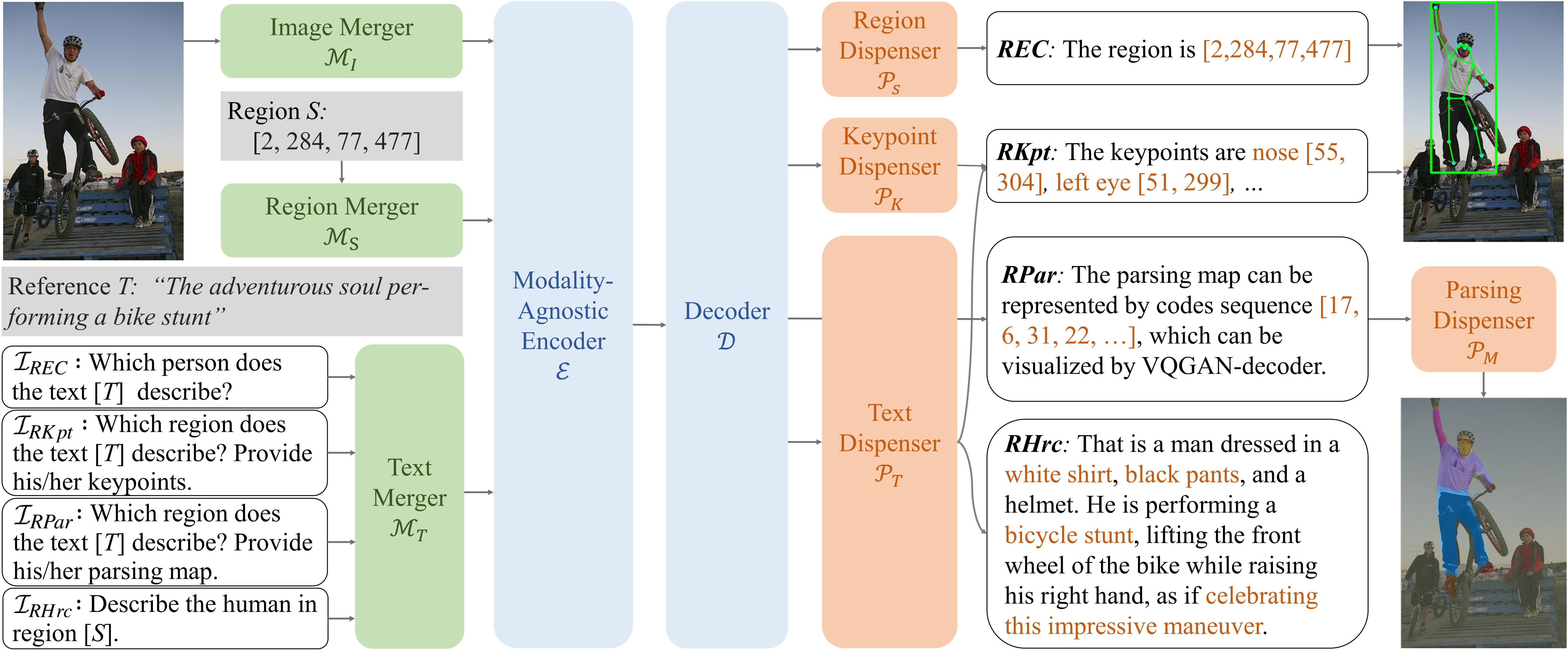}
    \caption{Overview of RefHCM model. RefHCM can handle four referring human-centric tasks in a unified way. Taking the referring keypoint task as an example, RefHCM first tokenizes the input image using the image merger $\mathcal{M}_I$ and the corresponding task instruction using the text merger $\mathcal{M}_T$. The resulting token sequence is then passed through 
 the encoder-decoder architecture to generate the desired output token sequence. Finally, the keypoint dispenser $\mathcal{P}_K$ transforms the output tokens into human keypoints.} 
    \label{fig:arch}
\end{figure*}

In this work, we aim to develop a unified \textit{referring} human-centric model that predicts human  perceptions of a referred individual using user-friendly text.
To achieve this, we focus on four referring human-centric tasks that integrate referring text into traditional human-centric tasks: localization, pose estimation, parsing, and captioning (including attribute recognition). Examples of these tasks are illustrated in Fig. \ref{fig:arch}.

\vspace{0.4em}

\noindent
\textbf{Referring Expression Comprehension (\emph{REC}).} \
\emph{REC} involves locating the region of a target person based on textural description. The instruction for \emph{REC} task is defined by the referring text $T$ and a task-specific instruction template $\mathcal{I}_{REC}$, \textit{e.g.}, ``Which person does the text [$T$] describe?''. Formally, given an image $I$  and  the referring-text instruction $\mathcal{I}_{REC}\left(T\right)$, \emph{REC} requires  predicting the region $S$ of the target person. Denote the unified referring model as $F$, \emph{REC} task can be formulated as:
\begin{equation}
 F\left(I, \mathcal{I}_{REC}\left(T\right)\right) \rightarrow S.
 \end{equation} 

\vspace{0.4em}

\noindent
\textbf{Referring Keypoint (\emph{RKpt}).} \
\emph{RKpt} involves predicting keypoints for individuals specified by input descriptions. The task instruction is determined by the referring text $T$ and instruction template $\mathcal{I}_{RKpt}$, such as ``Which region does the text [$T$] describe? Provide his/her keypoints''. Formally, given an image $I$  and instruction $\mathcal{I}_{RKpt}\left(T\right)$,  \emph{RKpt} aims to predict the keypoint $K\in\mathbb{R}^{N\times2}$ of the referred person, where $N$ is the number of keypoints. This task is formulated as:
\begin{equation}
 F\left(I, \mathcal{I}_{RKpt}\left(T\right)\right) \rightarrow K.
 \end{equation} 

\vspace{0.4em}

\noindent
\textbf{Referring Parsing (\emph{RPar}).} \
\emph{RPar} focuses on predicting parsing map for individuals specified by input descriptions. The instruction of \emph{RPar} is determined by the referring text $T$ and instruction template $\mathcal{I}_{RPar}$, \textit{e.g.}, ``Which region does the text [$T$] describe? Provide his/her parsing map''. Formally, given an image $I$  and the instruction $\mathcal{I}_{RPar}\left(T\right)$,  \emph{RPar} predicts the parsing map $M\in\mathbb{R}^{H\times W \times L}$ of the referred person, where $L$ is the number of body parts, and $H$ and $W$ denote the spatial dimensions.  \emph{RPar} task is formulated as 
\begin{equation}
F\left(I, \mathcal{I}_{RPar}\left(T\right)\right) \rightarrow M.
\end{equation} 

\vspace{0.4em}

\noindent
\textbf{Referring Human-Related Caption (\emph{RHrc}).} \
\emph{RHrc} focuses on providing detailed human-centric captions for specified visual regions $S$. The instruction for \emph{RHrc} is defined by the referring region $S$ and a template $\mathcal{I}_{RHrc}$, such as ``Describe the human in region [$S$]''. Formally, given an image $I$ and the instruction $\mathcal{I}_{RHrc}\left(S\right)$, the model is required to generate a comprehensive caption $T$ that describes the individual covering various aspects such as attire and activities. This task is formulated as 
\begin{equation}
F\left(I, \mathcal{I}_{RHrc}\left(S\right)\right) \rightarrow T.
\end{equation}

\section{RefHCM: A Unified Referring Human-centric Model}
\subsection{Overall}
We propose RefHCM, a general referring human-centric model, which can tackle various referring human-centric tasks in one unified model without any task-specific adaptation.
As shown in  Fig. \ref{fig:arch}, RefHCM consists of three key components: a sequence merger/dispenser responsible for building a unified representation space across various human-related modalities (Sec. \ref{sec3.2}), and a modality-agnostic encoder and decoder that process various human-centric referring tasks via a unified sequence-to-sequence paradigm (Sec. \ref{sec3.3}). To effectively address these referring tasks, we employ a three-stage training scheme encompassing sequence merger/dispenser training, referring captioning pre-training and multi-task joint training (Sec. \ref{sec3.4}). 
Overall, given an input image and instruction, RefHCM generates results through a three-step pipeline:

\textit{Step1: Merge modality-specific input into a unified sequence.} Given input image $I$ and task-specific instruction $\mathcal{I}$ from one of four referring tasks, we leverage corresponding merger $\mathcal{M}$ to project input data into a token sequence  $\mathbf{p}$. It can be formulated as $\mathbf{p}=\mathrm{concat}\left(\mathcal{M}_\mathrm{I}\left(I\right), \mathcal{M}_\mathrm{T}\left(\mathcal{I}\right)\right)$, where $\mathcal{M}_\mathrm{m}$ denotes the merger of modality $\mathrm{m} \in \left\{I,T\right\}$ and $\mathrm{concat}$ is the concatenation operation along the sequence dimension.

\textit{Step2: Generate modality-agnostic sequence with universal encoder-decoder.} RefHCM is build on encoder-decoder architecture. Given the input token sequence $\mathbf{p}$, the encoder $\mathcal{E}$ is employed to extract human-centric representations. These representations are then translated into desired output token sequence $\mathbf{q}$ by the output decoder $\mathcal{D}$ in an autoregressive manner. This sequence-to-sequence paradigm is formulated  as $\mathbf{q}=\mathcal{D}\left(\mathcal{E}\left(\mathbf{p}\right)\right)$.

\textit{Step3: Decode the output sequence into corresponding modality and optimize by a unified loss function.}
    Given the output tokens $\mathbf{q}$ after the decoder, the designated output modality can be generated by the modality-specific sequence dispenser $\mathcal{P}$. This process can be formulated as $\hat{\mathbf{y}}_\mathrm{m}=\mathcal{P}_\mathrm{m}\left(\mathbf{q}\right)$ where $\mathcal{P}_\mathrm{m}$ denotes the dispenser of modality $\mathrm{m}$ with $\mathrm{m} \in \left\{T, S, K, M\right\}$. 
    Finally, the overall parameters of RefHCM are optimized by a unified autoregressive loss.

\subsection{Sequence Merger and Dispenser}
\label{sec3.2}
A primary challenge in developing a \textit{unified} referring human-centric model is to unify heterogeneous inputs and outputs modalities. Previous approaches typically rely on modality-specific projectors and task-specific heads to manage diverse inputs and outputs. Differently, RefHCM employs sequence mergers and dispensers, effectively unifying these heterogeneous modalities into a shared representation space.

\vspace{0.4em}

\noindent
\textbf{Image.} \
We employ ResNet152 \cite{kaiming2016resnet} as the image merger $\mathcal{M}_\mathrm{I}$, which  converts the image into a sequence of patch features. These patch features are then flattened and aligned with the text embeddings through a linear resampler.
Formally, given an input image $I$, the image merger $\mathcal{M}_\mathrm{I}$ generates the image tokens $\mathbf{p}_\mathrm{I}$ as follows:
\begin{equation} 
\label{eq:merger_i} 
\begin{aligned} 
\mathbf{p}_\mathrm{I} = \mathcal{M}_\mathrm{I}\left(\textit{I}\right) = \mathrm{Resampler}\left(\mathrm{Flatten}\left(\mathrm{ResNet}\left(\textit{I}\right)\right)\right). 
\end{aligned} 
\end{equation} 

\vspace{0.4em}

\noindent
\textbf{Text.} \
Following GPT \cite{openai2023gpt4} and BART \cite{mike2020bart}. we apply Byte Pair Encoding (BPE) algorithm to transform text  into subword sequence, which  are subsequently mapped into a sequence of tokens. Formally, given a text sequence $T$, the text merger $\mathcal{M}_\mathrm{T}$ generates the text tokens $\mathbf{p}_{\mathrm{T}}$ as follows:
\begin{equation}
\mathbf{p}_{\mathrm{T}}=\mathcal{M}_\mathrm{T}\left(T\right)=\mathrm{BPE}\left(T\right).
\end{equation}
Similarly, given output text tokens, we can recover the output text using the inverse BPE mapping.

\vspace{0.4em}

\noindent
\textbf{Spatial Coordinates.} \
Spatial information plays a pivotal role  in human perception. For example, both \emph{REC} and \emph{RKpt} demand  the prediction of spatial coordinates, where \emph{REC} focuses on identifying the human bounding box while \emph{RKpt} targets  human keypoints.  The coordinate merger $\mathcal{M}_\mathrm{C}$ and dispenser $\mathcal{P}_\mathrm{C}$ are implemented using spatial quantization and dequantization operations. Formally, denote a corner coordinate as $\left(x, y\right)$, 
$\mathcal{M}_\mathrm{C}$ uniformly discretizes  this continuous corner coordinate  into integer, producing bin tokens:
\begin{equation}
\label{eq:spatial_quantization}
\begin{aligned}
\mathbf{p}^{x}_{\mathrm{C}}&=\mathcal{M}_\mathrm{C}\left(x\right)=\lfloor \frac{x}{W} \times N \rfloor, \\
\mathbf{p}^{y}_{\mathrm{C}}&=\mathcal{M}_\mathrm{C}\left(y\right)=\lfloor \frac{y}{H} \times N \rfloor,
\end{aligned}
\end{equation}where $N$ is a hyper-parameter to control the quantization precision, $W$ and $H$ denote the width and height of input image, respectively, and $\lfloor . \rfloor$ is floor operation.
$\mathcal{P}_\mathrm{S}$ maps the location tokens back to coordinates in original image space, serving as the inverse process of $\mathcal{M}_\mathrm{C}$. Formally, given bin tokens $\mathbf{q}_C$ output by the model, $\mathcal{P}_\mathrm{C}$ obtains the output coordinates as:
\begin{equation}
\begin{aligned}
\hat{\mathbf{y}}^{x}_{\mathrm{C}}&=\mathcal{P}_\mathrm{C}\left(\mathbf{q}^{x}_{\mathrm{C}}\right)=\lceil \frac{\mathbf{q}^{x}_{\mathrm{C}}}{N} \times W \rceil, \\
\hat{\mathbf{y}}^{y}_{\mathrm{C}}&=\mathcal{P}_\mathrm{C}\left(\mathbf{q}^{y}_{\mathrm{C}}\right)=\lceil \frac{\mathbf{q}^{y}_{\mathrm{C}}}{H} \times N \rceil,
\end{aligned}
\end{equation}
where $\lceil . \rceil$ is ceil operation.

\textit{\textbf{Region.}} \
We represent regions within an image using bounding boxes. Specifically, Each region $S$ is defined by two points $\left(x_S^1, y_S^1\right)$ and $\left(x_S^2, y_S^2\right)$, corresponding to the top-left and bottom-right corners of the box respectively. The region merger $\mathcal{M}_\mathrm{S}$ converts these two spatial coordinates into corresponding bin tokens using $\mathcal{M}_\mathrm{C}$, while the region dispenser $\mathcal{P}_\mathrm{S}$  performs the inverse operation, converting generated position tokens back into coordinates using $\mathcal{P}_\mathrm{C}$.

\textit{\textbf{Keypoints.}} \ 
A straightforward approach is to represent keypoints as a series of coordinates. However, this representation lacks semantic clues for each keypoint, 
which increases the difficulty of model learning. To address this, we propose integrating semantic information for each keypoint, such as its name (\textit{e.g.}, nose, wrist). 
Thus, each keypoint is represented by both its name and the corresponding coordinates. Formally, given the human keypoints $K\in\mathbb{R}^{N\times2}$, the keypoint merger $\mathcal{M}_{\mathrm{K}}$ and dispenser $\mathcal{P}_{\mathrm{K}}$ are formulated as:
\begin{equation}
\begin{aligned}
\mathbf{p}_{\mathrm{K}} &= \mathcal{M}_{\mathrm{K}}(K) = \{\mathrm{Name}_i, \mathcal{M}_{\mathrm{C}}(x_K^i), \mathcal{M}_{\mathrm{C}}(y_K^i)\}_{i=1}^N, \\
\hat{\mathbf{y}}_\mathrm{K} &= \mathcal{P}_\mathrm{K}\left(\mathbf{q}_\mathrm{K}\right) = \{\mathcal{P}_\mathrm{C}\left(\mathbf{q}_\mathrm{K}^i\right)\}_{i=1}^N,
\end{aligned}
\end{equation}where
$\mathrm{Name}_i$, $\left(x_K^i,  y_K^i\right)$ and $\mathbf{q}_\mathrm{K}^i$ denote the name, coordinate and predicted token for the $i$-th keypoint.
This keypoint representation provides additional semantic supervision signals for pose estimation. Moreover, the use of name-coordinate pairs
introduces a more flexible sequential representation, eliminating the need to predict a fixed set of points and  visibility flags, as required in traditional pose estimation methods.

\begin{figure}[t]
    \centering
    \includegraphics[width=0.5\textwidth]{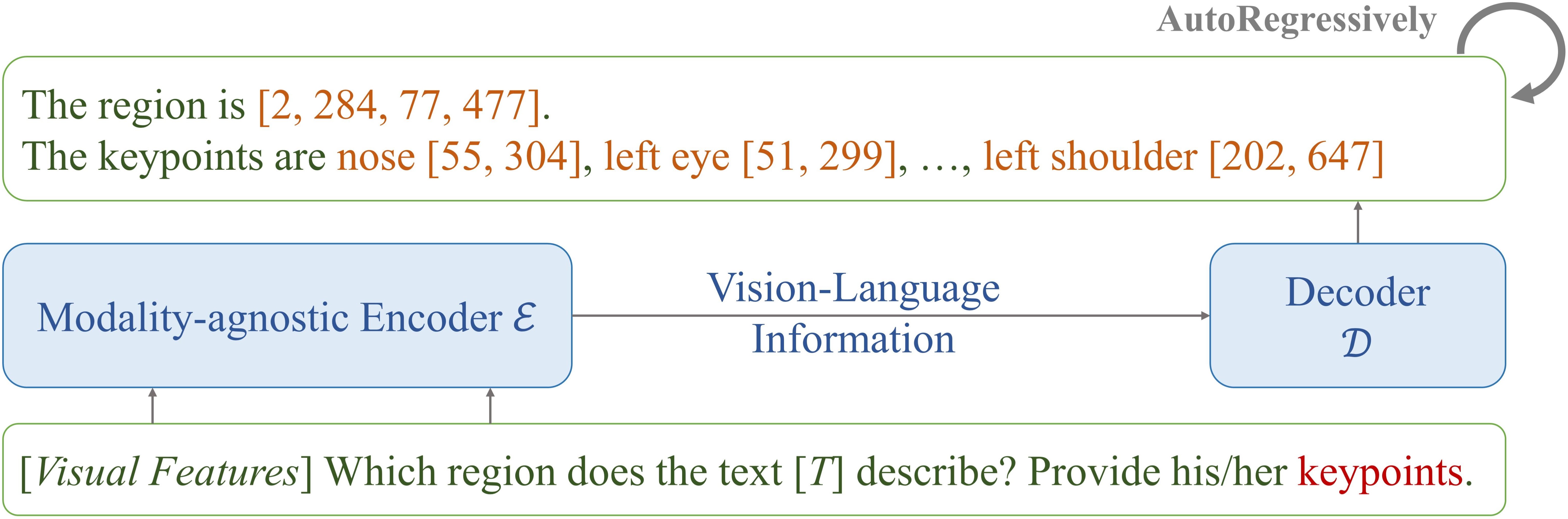}
    \caption{The illustration of Location-Context Restriction, which prepends the human bounding box to the keypoint output of \emph{RKpt} task. Through the autoregressive decoding process, bounding boxes and keypoints are generated sequentially, allowing for mutual positive constraints between them.}
    \label{fig:refpose-form}
\end{figure}

\textit{\textbf{Location-Context Restriction.}} 
Both \emph{REC} and \emph{RKpt} tasks require the output of spatial coordinates: \emph{REC} necessitates  the person's bounding box, while \emph{RKpt} requires the person's keypoints. 
Intuitively, keypoint information can assist in localizing the bounding boxes, as an ideal bounding box should encompass all keypoints. Building on this insight, we propose  the Location-Context Restriction (LCR), which prepends the human bounding box to the output of \emph{RKpt} task.  This approach is analogous to the "Chain of Thought" approach in Natural Language Processing \cite{jason2022cot}. As illustrated in Fig. \ref{fig:refpose-form}, instructions containing the term \textit{keypoints} prompt the model to prioritize the keypoint information of the referred person, thereby improving the generation of bounding box coordinates while concurrently constraining the generation of keypoints.
In essence, by integrating human-related location information from both \emph{REC} and \emph{RKpt} tasks, LCR enhances the performance of both tasks, highlighting their inherent homogeneity.

\vspace{0.4em}

\noindent
\textbf{Parsing Map.} \
To represent human parsing map in discrete semantic tokens, we build the parsing merger and dispenser based on Vector Quantized Variational Autoencoders (VQ-VAE). 
The VQ-VAE architecture consists of a parsing encoder $\mathcal{E}_m$, a parsing decoder $\mathcal{D}_m$, along with a codebook $\mathcal{B} = \left\{b^1, \dots, b^{N_m}\right\}$  containing $N_m$ embeddings. 
Formally, given a parsing map $M\in\mathbb{R}^{H\times W \times L}$, where each channel corresponds  to a body part and the pixel values are binary indicating whether a pixel belongs to corresponding body part, the parsing encoder $\mathcal{E}_m$ that consists of several  2-D convolutional layers projects $M$ to a latent embeddings $z\in\mathbb{R}^{h\times w \times d}$. Here, $h$ and $w$ are the height and width after downsampling and $d$ is the latent dimension. Next, we transform $z$ into a collection of codebook entries via discrete quantization. Specifically, the quantization process replaces each item of z with the nearest embedding in the codebook $\mathcal{B}$, producing the quantized latent vectors $\hat{z} \in \mathbb{R}^{h\times w \times d}$ as follows:
\begin{equation}
\hat{z} = \mathop{\arg\min}_{b^k\in \mathcal{B}}\left\|z-b^k\right\|_2.
\end{equation}
The parsing decoder $\mathcal{D}_m$, which consists of several 2-D deconvolutional layers, project the quantized embedding back to raw parsing-map space, \textit{i.e.}, $\hat{M}=\mathcal{D}_m\left(\hat{z}\right)$. We train the VQ-VAE by a pixel-wise classification loss, embedding loss and commitment loss as follows:
\begin{equation}
\begin{aligned}
\label{eq_vqvae}
\mathcal{L}_{\mathrm{vq}} = &-M\log\left(\hat{M}\right) - \left(1-M\right) \log \left(1-\hat{M}\right) \\
&+ \left\|\mathrm{sg}\left[z\right]-\hat{z}\right\| + \beta\left\|\mathrm{sg}\left[\hat{z}\right]-z\right\|,
\end{aligned}
\end{equation}
where $\mathrm{sg}\left[\cdot\right]$ is the stop gradient operation, and $\beta$ is the coefficient to adjust the weight of the commitment loss. 

After training the VQ-VAE, we can instantiate the parsing merger $\mathcal{M}_{\mathrm{M}}$ and dispenser $\mathcal{P}_{\mathrm{M}}$ based on the trained parsing encoder $\mathcal{E}_m$ and decoder $\mathcal{D}_m$, respectively. Specifically,  $\mathcal{M}_{\mathrm{M}}$ quantizes a parsing map $M$ into a sequence of discrete codebook-indices of quantized embedding vector, namely parsing tokens $\mathbf{p}_{\mathrm{M}}$, as follows:
\begin{equation}
\mathbf{p}_{\mathrm{M}} = \mathcal{M}_{\mathrm{M}}\left(M\right)=\mathop{\arg\min}_{k\in\left\{1, \dots, N_m\right\}}\left\|\mathcal{E}_m\left(M\right)-b^k\right\|_2.
\end{equation}
During the inference process, the parsing tokens are decoded back into their original space by parsing dispenser $\mathcal{P}_{\mathrm{M}}$. Formally, given the parsing tokens $\mathbf{q}_\mathrm{M}$ predicted by the model, $\mathcal{P}_{\mathrm{M}}$ obtains the output parsing map as:
\begin{equation}
\hat{\mathbf{y}}_{\mathrm{M}}=\mathcal{P}_{\mathrm{M}}\left(\mathbf{q}_\mathrm{M}\right)=\mathcal{D}_m\left(\mathcal{B}\left[\mathbf{q}_\mathrm{M}\right]\right),
\end{equation}
where $\mathcal{B}\left[\mathbf{q}_\mathrm{M}\right]$ represent the $\mathbf{q}_\mathrm{M}$-th entry in codebook $\mathcal{B}$.

\subsection{Universal Encoder and Decoder}
\label{sec3.3}
Following previous works, we employ a Transformer-based backbone architecture, and adopt a unified encoder-decoder framework to address all human-centric referring tasks.  
Specifically, given the input token sequence $\mathbf{p}$ extracted by the sequence merger, the encoder $\mathcal{E}$ processes the input sequence into human-centric representations, denoted as $\mathcal{E}\left(\mathbf{p}\right)$. 
These representations are subsequently passed through the decoder $\mathcal{D}$, which generates the output token sequence $\mathbf{q}=\mathcal{D}\left(\mathcal{E}\left(\mathbf{p}\right)\right)$.  
Finally, the output tokens are decoded back into their original representation by associated Dispenser.
Both encoder and decoder are composed of stacked Transformer layers. Each Transformer encoder layer consists of a  self-attention and a feed-forward network (FFN).  Each Transformer decoder layer incorporates a self-attention, FFN and a cross attention layer to establish connection between the encoder and decoder representations. 

\begin{figure}[t]
    \centering
    \includegraphics[width=0.5\textwidth]{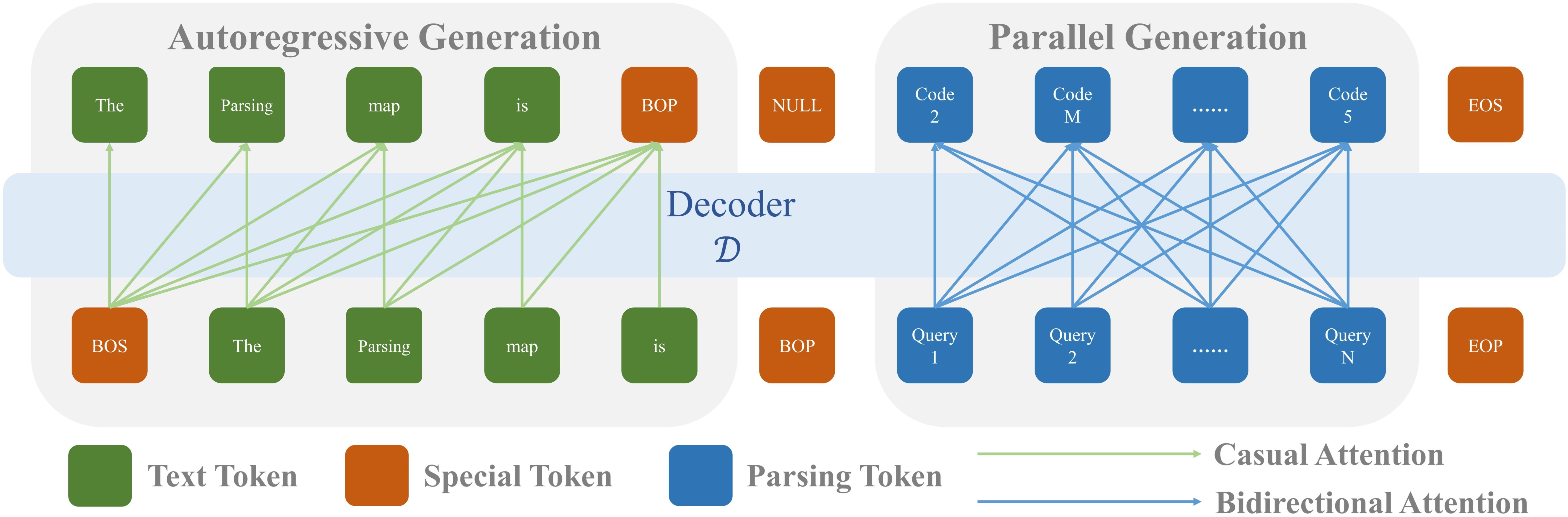}
    \caption{Overview of QPG (Query Parallel Generation), which significantly boosts inference speed. It is worth noting that Queries can see each other, akin to full mask attention. During the inference phase, the generation method shifts from auto-regressive to parallel generation upon encountering the parsing map query token $<\text{BOP}>$. M represents the size of codebook in the parsing VQ-VAE, which also corresponds to the prediction range for the decoder.}
    \label{fig:qpg}
\end{figure}

Notably, the decoder generates the target sequence in an autoregressive manner. However, for certain modalities, such as parsing maps with a large number of parsing tokens, the efficiency of autoregressive manner is significantly limited. Additionally, unlike the inherently sequential nature of text tokens, parsing-map tokens—encoding spatial semantics of human body parts—posses a fundamentally non-sequential structure. This property makes them particularly well-suited for parallel generation, which offers a potential avenue for improving efficiency while maintaining structural coherence. Inspired by \cite{transfusion2024}, we modify the standard causal attention in the transformer decoder by incorporating  bidirectional attention. 
As shown in Fig.~\ref{fig:qpg}, we propose QPG (Query Parallel Generation), which applies bidirectional attention and parallel generation to parsing-map token sequence while maintaining  casual attention and autoregressive generation to token sequences of other modalities. 
Specifically, we define $N = h\times w$ learnable queries, denoted as $\mathcal{Q}=\left\{<\mathrm{Query}_1>, \dots, <\mathrm{Query}_N> \right\}$, which are used to predict corresponding parsing-map tokens in a single forward step.
To establish a seamless connection between the encoder-decoder architecture and the parsing VQ-VAE, these learnable queries are initialized using the codebook of the parsing VQ-VAE. This initialization enables the encoder-decoder to effectively interpret the parsing map rather than `blindly'  generating parsing tokens, enhancing parsing prediction ability.

\subsection{Optimization}
\label{sec3.4}

\noindent
\textbf{Objective Function.} 
With the integration of a sequence merger/dispenser and encoder-decoder architecture, the different human-centric tasks can be formulated within a sequence-to-sequence paradigm.
This formulation enables the unification of diverse task-specific losses into a single classification loss, where the predicted tokens are supervised by corresponding ground truth tokens.  
Specifically, given the predicted token sequence $\hat{\mathbf{y}}$ generated by RefHCM and the ground truth token sequence $\mathbf{y}$, the cross-entropy classification loss is computed as follows:
\begin{equation} \begin{aligned} 
\mathcal{L}_{CE} = - \sum_{i=1}^{\vert \mathbf{y} \vert} \mathbf{y_i} \log P_\theta (\hat{\mathbf{y}_i} \vert \hat{\mathbf{y}}_{\textless i}),
\end{aligned} \end{equation}where $\theta$ refers to the model parameters.
While incorporating task-specific losses (\textit{e.g.}, reconstruction loss for parsing maps) can improve performance, we prioritize simplicity and efficiency. Therefore, we rely solely on the cross-entropy classification loss,  effectively addressing GPU memory constraints while maintaining a unified architecture framework.

\vspace{0.4em}

\noindent
\textbf{Training Paradigm}.
The training process of RefHCM is divided into three stages.  

\textit{Stage1: Sequence Merger/Dispenser Training.} In the first stage, we train the sequence merger/dispenser. Notably, only the parsing merger/dispenser requires parameter optimization, while the other mergers/dispensers are implemented with fixed configurations. So we  train only the parsing  merger/dispenser  using the objective defined in Eq. \ref{eq_vqvae}. These sequence mergers and dispensers allows any human-related modality to be represented as a sequence of tokens, enabling seamless integration within a sequence-to-sequence framework.

\textit{Stage2:} \emph{RHrc} \textit{Pre-Training.} After learning sequence mergers and dispensers, we proceed to pre-train the encoder-decoder architecture using the \emph{RHrc} task. 
 \emph{RHrc} involves generating detailed captions that describe persons, which assist in analyzing human attributes for other human-centric tasks.
This pre-training stage ensures that the model develops strong language comprehension and expression capabilities in relation to human representation.

\textit{Stage3: Multi-task Training.} Finally, we combine all four human-centric referring tasks for training. In implementation, to maintain the stability of multi-task training, we ensure that all task appear within a single batch. Additionally, we increase the sampling probability of the \emph{RPar} task, as generating parsing map is relatively more challenging.

\begin{figure}[t]
    \centering
    \includegraphics[width=0.5\textwidth]{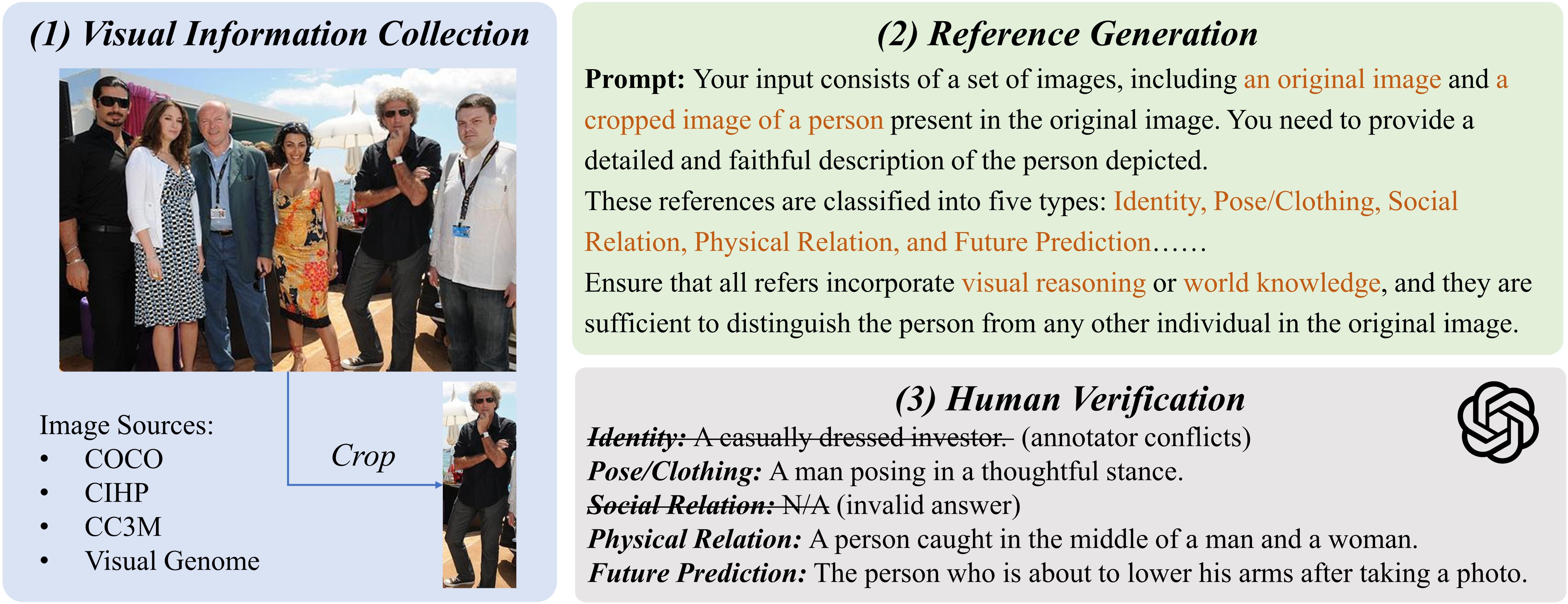}
    \caption{\emph{ReasonRef} benchmark construction pipeline. We use GPT-4 to generate descriptions across five dimensions covering identity, pose/clothing, relations, and future prediction, then manually verify generated descriptions.}
    \label{fig:reasonref_dec}
\end{figure}

\section{Reason Reference Benchmark}
\label{sec3.5}
To further evaluate models' ability to predict visual perceptions through \textit{implicit} references requiring \textit{complex reasoning}, we introduce the Reasoning Reference Benchmark (\emph{ReasonRef}). This benchmark consists of three task-specific subsets: \emph{ReasonDet}, \emph{ReasonKpt}, and \emph{ReasonPar}, which focus on predicting regions, keypoints and parsing maps for individuals specified by reasoning references.  Additionally, to assess model's reasoning capabilities under finetuning conditions, we have curated a training set \emph{$\text{ReasonRef}_{\text{train}}$}. The overall framework for benchmark construction is shown in Fig. \ref{fig:reasonref_dec}.  

\vspace{0.4em}

\noindent
\textbf{Task Dimensions.} \
\label{reasonref_task_dimension}
To comprehensively assess  human-centric reasoning capabilities, \emph{ReasonRef} incorporates five evaluation dimensions covering identity, pose/clothing, relations, and future prediction, as outlined below:
\textbf{}
\begin{itemize}
    \item \textbf{Identity.} A complex reasoning reference regarding the person’s identity, such as their occupation based on the observation of their appearance and clothing.
    \item \textbf{Pose/Clothing.} A complex reasoning reference related to the person’s body posture or dress style, including their body orientation and specific attire details.  
    \item \textbf{Social Relations.} A complex reasoning reference concerning the social relationships  among multiple individuals, encompassing their roles within the environment.
    \item \textbf{Physical Relations.} A complex reasoning reference to spatial relationships between an individual and other people or objects,  including their relative positions, movements, and physical connections within the scene.
    \item \textbf{Future prediction.}  A predictive reference regarding the person’s likely next action or movement, based on observed dynamics in the scene.
\end{itemize}

\vspace{0.2em}

\noindent
\textbf{Benchmark Construction.} \
As shown in Fig. \ref{fig:reasonref_dec}, we employ a GPT-assisted pipeline for benchmark construction, involving both reference generation and verification. 
\textbf{\textit{(1) Visual Information Collection.}} To gather diverse visual information for reference generation, we utilize a variety of image sources, including COCO \cite{COCO}, CIHP \cite{kegong2018cihp}, CC3M \cite{cc3m}, and Visual Genome \cite{visualgenome}.
These datasets encompass a board spectrum of visual scenarios, providing rich and varied data for generating referential text.
\textbf{\textit{(2) Reference Generation.}} We prompt GPT-4 to generate reasoning reference for each evaluation dimension. Specifically, we develop tailored prompts for each task dimension and provide both the entire image and a cropped target-person image to GPT-4. The entire image offers global contextual information, while the cropped person image focuses on local details.  By presenting these two perspectives, we ensure that GPT-4 can generate reference that require reasoning across both broad contextual understanding and fine-grained, person-specific details. 
\textbf{\textit{(3) Human Verification.}} To ensure the quality of \emph{ReasonRef}, we employ human annotators to verify the generated reference. Annotators are tasked with matching each reference with corresponding person, and any reference that conflicts with the ground-truth person or deviates from the intended task dimension is discarded.  

\vspace{0.4em}

\noindent
\textbf{Data Statistics.} \ 
As detailed in Table \ref{tab:reasonref_statis}, \emph{ReasonRef} consists of 6,551 expressions spanning 1,390 human instances. 
Each instance is annotated with up to 5 types of reasoning references. The benchmark is divided into three task-specific subsets: \emph{ReasonDet} with 472 instances along with bounding box annotations, \emph{ReasonKpt} with 464 instances along with keypoint annotations, and \emph{ReasonPar} with 454 instances along with parsing map annotations.
In contrast to previous benchmarks, we annotate only one person per image, the one with the largest area or the most complete keypoint and parsing annotations, to reduce the complexity of reference generation.

\begin{table*}[t]
\caption{Data statistics of human-related images in RefCOCO testA \cite{refcoco}, RefCOCO+ testA, and our \emph{ReasonRef}.}
\centering
\begin{tabular}{l|c|c|c|c}
\hline
{\textbf{Benchmark}} & {\textbf{Target}} & {\textbf{Image}} & {\textbf{Instance}} & {\textbf{Expression}} \\
\hline
RefCOCO testA & Boxes & 750 & 1975 & 5657 \\
RefCOCO+ testA & Boxes & 750 & 1975 & 5726 \\
\textbf{\emph{ReasonRef}} & \textbf{Boxes, Keypoints, Parsing maps} & \textbf{1390} & 1390 & \textbf{6551} \\
\hline
\end{tabular}
\label{tab:reasonref_statis}
\end{table*}

%% file: 4_experiments.tex
\section{Experiments}

\begin{table}[t]
\caption{Comparisons on \emph{REC} task. We report AP@50 metric on RefCOCO testA and RefCOCO+ testA datasets.}
\centering
\begin{tabular}{l|c|c|c}
\hline
{\textbf{Model}} & {\textbf{Model Size}} & {\textbf{RefCOCO A}} & {\textbf{RefCOCO+ A}} \\
\hline
VILLA  \cite{zhegan2020villa} & - &  87.48  & 81.54  \\
MDTETR \cite{ziyi2022mdetr} & 400M &  89.58 & 84.09  \\
OFA-L-MT \cite{pengwang2022ofa} & 520M & 80.84 & 76.15  \\
OFA-L-tuned \cite{pengwang2022ofa} & 520M & 92.93 &  \textbf{89.87} \\
UNITER \cite{yen2020uniter} & 870M &  87.04 &  81.45 \\
UNINEXT-H \cite{fangjian2023uninext} & 1B &  \textbf{94.33} &  \underline{89.63} \\
\hline
\textbf{RefHCM} & 500M & \underline{93.69} & 89.56 \\
\hline
\end{tabular}
\label{tab:refcoco-results}
\end{table}

\subsection{Datasets and Evaluation Metric}
To systematically evaluate human-centric referring tasks, we construct training and test datasets tailored to their specific requirements. 

\vspace{0.4em}

\noindent
\textbf{\emph{REC} Datasets}. 
We build  a new dataset based on the public RefCOCO series: RefCOCO \cite{refcoco}, RefCOCO+ \cite{refcoco}, and RefCOCOg \cite{refcocog}, each containing references of varying complexity. Given our focus on human perception, we curate the training data by retaining only human-related annotations. For a fair comparison, we evaluate performance on both  \textbf{RefCOCO testA} and \textbf{RefCOCO+ testA} subsets,  as these official test subsets  exclusively  consist of human-related annotations.

\vspace{0.4em}

\noindent
\textbf{\emph{RKpt} Datasets.}  
Building on the connection between RefCOCO series and COCO dataset, where RefCOCO images originate  from COCO \cite{COCO}, we integrate COCO's keypoint annotations with referring expressions. This integration gives rise to Refpose series, comprising   \textbf{Refpose}, \textbf{Refpose+} and \textbf{Refposeg}, derived from RefCOCO, RefCOCO+, and RefCOCOg, respectively.

\vspace{0.4em}

\noindent
\textbf{\emph{RPar} Datasets.}  
We select the CIHP dataset \cite{kegong2018cihp} as the source for parsing map annotations. And we generate reference text for each individual in the CIHP dataset using both Ferret-13B \cite{haoxuanyou2024ferret} and GPT-4V \cite{openai2023gpt4}, forming the \textbf{RefCIHP} dataset. 
For the test set, eferences are exclusively generated by GPT-4V and subsequently verified through  human review. 

\vspace{0.4em}

\noindent
\textbf{\emph{RHrc} Datasets.}  
Previous approaches often rely on the reversed RefCOCO series as region captions, but these descriptions frequently fall short in providing the depth and breadth needed for comprehensive human descriptions. To address this limitation, we leverage the advanced capabilities of MLLM (Ferret-13B \cite{haoxuanyou2024ferret}) to generate detailed, multi-aspect captions for images in the CIHP dataset \cite{kegong2018cihp}. These enhanced captions provide rich information about individuals, including their appearance, attire, and activities. The resulting dataset, termed \textbf{CapCIHP}, comprises approximately 100k annotated instances.

\vspace{0.4em}

\noindent
\textbf{\emph{ReasonRef} Datasets.} 
This benchmark is specifically crafted to evaluate a model's ability to comprehend reasoning-based references. It includes three subsets: \emph{ReasonDet}, \emph{ReasonKpt}, and \emph{ReasonPar}, corresponding to \emph{REC}, \emph{RKpt}, and \emph{RPar} tasks, respectively.
Detailed descriptions and analysis are provided in Section \ref{sec3.5}

\vspace{0.4em}

\noindent
\textbf{Evaluation Metrics.} \
For the \emph{REC} task, we employ the AP@50 metric, which evaluates object detection precision by measuring the overlap between predicted bounding boxes and ground truth at a 50\% Intersection over Union (IoU) threshold.
For the \emph{RKpt} task, we use the Object Keypoint Similarity (OKS) AP metric \cite{COCO}, assessing keypoint localization accuracy by calculating the similarity between predicted and ground-truth keypoints, adjusted for object scale.
For the \emph{RPar} task, the mean IoU (mIoU) metric is utilized to evaluate parsing accuracy by averaging IoU scores across all classes, ensuring balanced performance.
For the \emph{RHrc} task, CIDEr \cite{rama2015cider} measures the quality of generated human-related captions by evaluating their naturalness and relevance.

\begin{table*}[htbp]
\begin{minipage}[t]{0.65\linewidth}
\caption{Comparisons on \emph{RKpt} task. We report AP@50 for bounding box prediction and OKS AP metric for keypoints prediction on Refpose series datasets.}
\label{tab:refpose-results}
\centering
\begin{tabular}{l|c|cc|cc|cc}
\hline
\multirow{2}{*}{\textbf{Model}} & \multirow{2}{*}{\textbf{Model Size}} & \multicolumn{2}{c|}{\textbf{Refpose}} & \multicolumn{2}{c|}{\textbf{Refpose+}} & \multicolumn{2}{c}{\textbf{Refposg}} \\
& & AP@50 & OKS AP & AP@50 & OKS AP & AP@50 & OKS AP \\
\hline
Unified-IO-2 \cite{jiasen2024uio2} & 1.1B & 89.13 & 52.00 & 82.35 & 48.25 & 89.94 & 54.01 \\
$PoseGPT_{text}$ \cite{yaofeng2023posegpt} & 500M & 78.7 & 70.50 & 82.03 & 71.46 & 91.94 & \textbf{76.77} \\
\hline
\textbf{RefHCM} & 500M & \textbf{93.69} & \textbf{75.60} & \textbf{89.56} & \textbf{72.24} & \textbf{93.42} & 75.69 \\
\hline
\end{tabular}
\end{minipage}
\hfill
\begin{minipage}[t]{0.3\linewidth}
\caption{Comparisons on \emph{RKpt} task. We report the OKS AP metric on RefHuman dataset.}
\vspace{1em}
\label{tab:refhuman-results}
\centering
\begin{tabular}{l|c|c}
\hline
{\textbf{Model}} & {\textbf{Model Size}} & {\textbf{OKS AP}} \\
\hline
UniPHD \cite{uniphd2024miao} & 184M & 66.7 \\
\hline
\textbf{RefHCM} & 500M & \textbf{66.8} \\
\hline
\end{tabular}
\end{minipage}
\end{table*}

 \begin{figure}[t]
    \centering
    \includegraphics[width=0.45\textwidth]{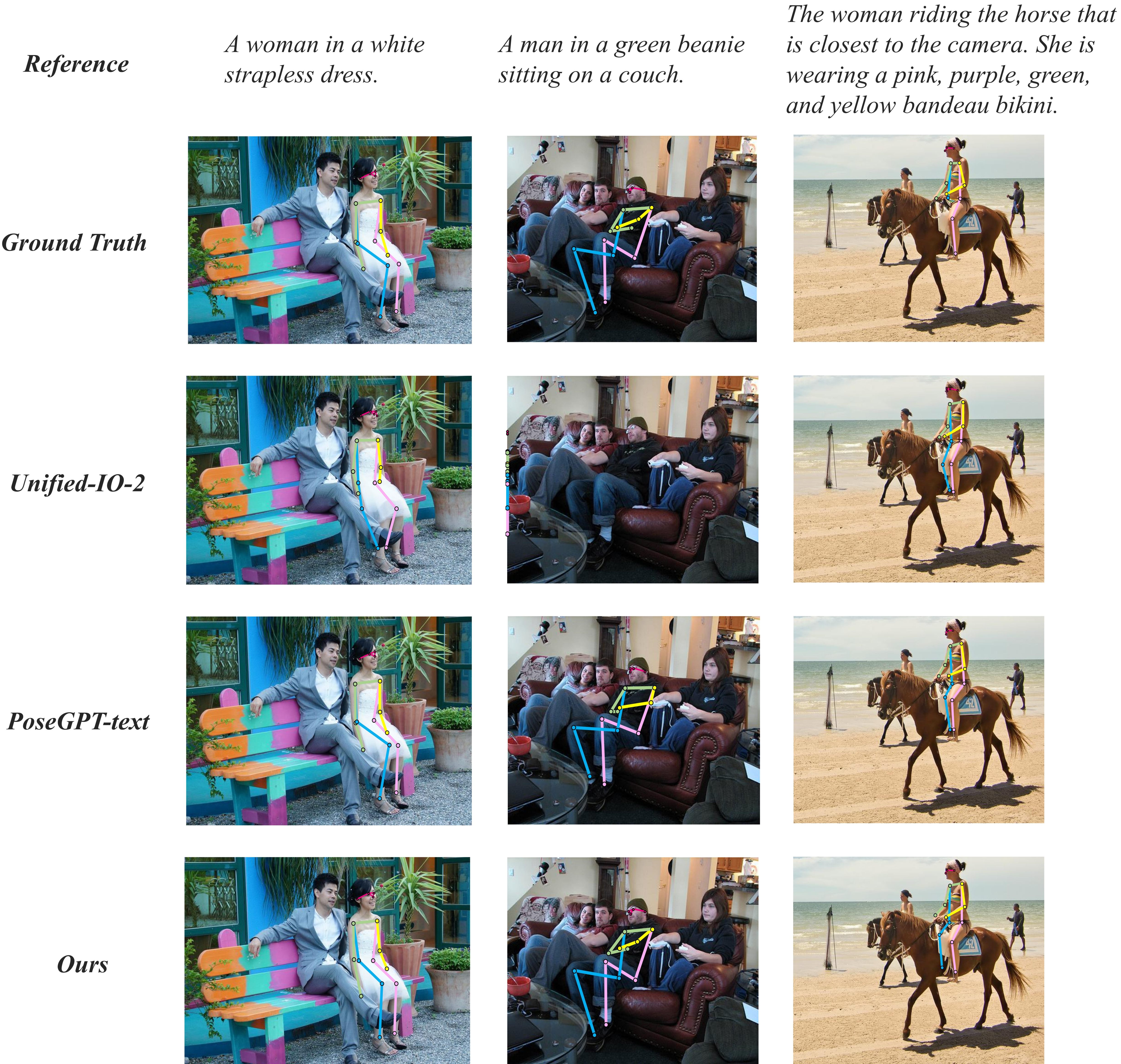}
    \caption{Qualitative results on \emph{RKpt} task. RefHCM produces more precise predictions for occluded keypoints.}
    \label{fig:rkpt-example}
\end{figure}

\begin{figure}[t]
    \centering
    \includegraphics[width=0.45\textwidth]{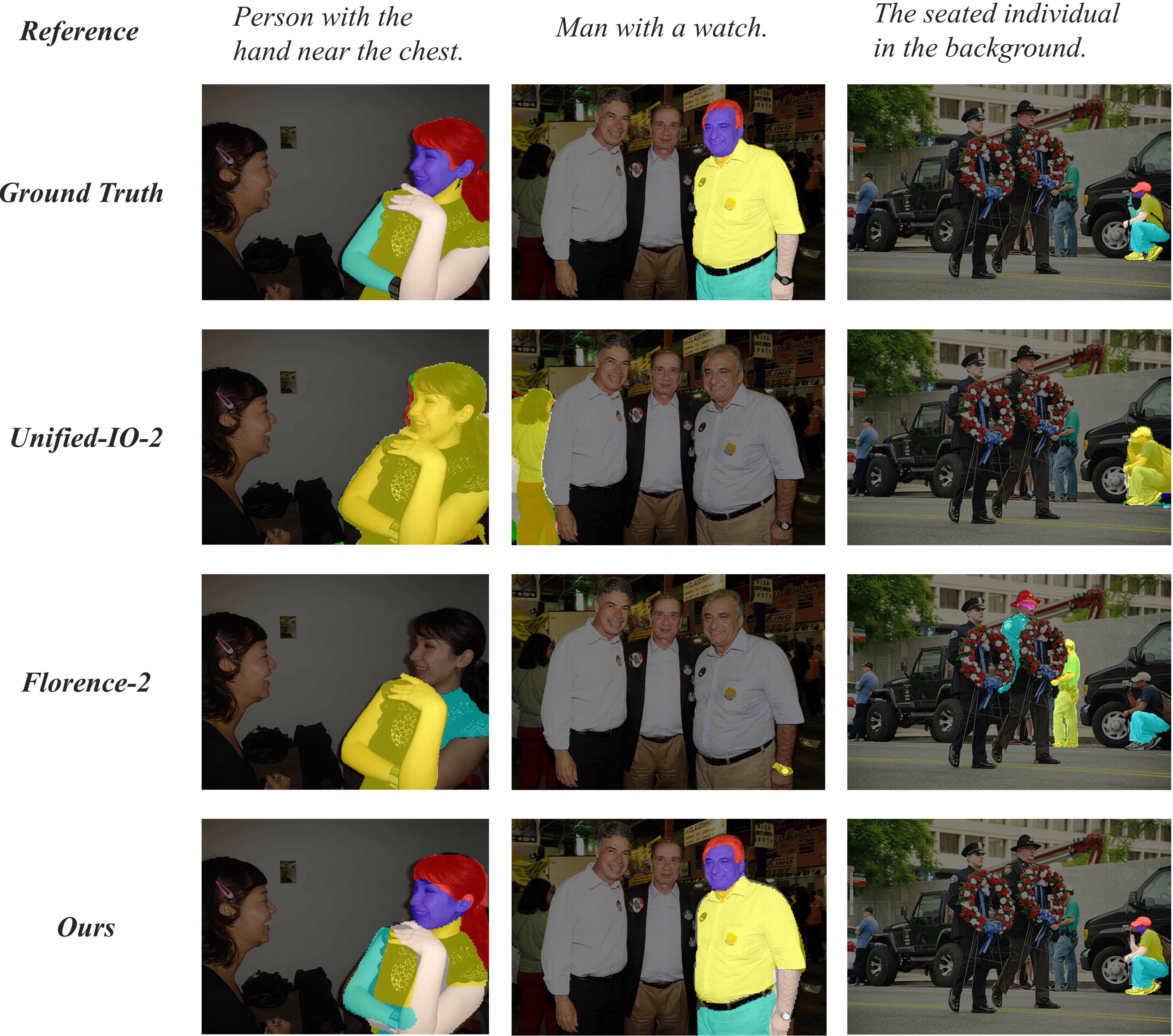}
    \caption{Qualitative results on \emph{RPar} task. RefHCM can more precisely locate the target human and generate his/her parsing map.}
    \label{fig:rpar-example}
\end{figure}

\subsection{Implementation Details}


\noindent
\textbf{Training of Parsing Merger/Dispenser.} \ 
We instantiate the parsing merger/dispenser using a \textbf{modified VQGAN} architecture   tailored for parsing maps rather than RGB images. Compared to original VQGAN \cite{patrick2021vqgan}, the modifications include: (1) \textit{Simplified Structure}: Give that parsing maps have a relatively simple structure, we reduce the codebook size from $8192$ to $32$ and increase the downsampling rate from $8$ to $16$. This adjustment effectively reduces the number of parsing tokens, simplifying the prediction task. (2) \textit{Enhanced Coherence}: Since human body parts typically exhibit coherence, we add incorporate two additional self-attention layers in the VQGAN encoder and decoder. This enhances the model's ability to aggregate relevant features and improve representation quality. 
(3) \textit{Aspect Ratio Optimization}: Since parsing maps represent individual subjects, retaining the original square resolution of VQGAN would introduce redundant space around the edges. To address this, we adjust the aspect ratio of the parsing map to $4:3$, derived from the bounding box statistics of individuals in the training data.

During training the parsing VQVAE, we set the codebook size to $32$ with a downsampling ratio of $16$. The input dimension of each parsing map is set to $128 \times 96$. 
Training is conducted using the Adam optimizer with a batch size of 20 and an initial learning rate of $4.5 \times 10 ^{-6}$. 

\vspace{0.4em}

\noindent
\textbf{Training of the encoder-decoder architecture.} \ 
To reduce the training cost of aligning text and visual modalities, we partially initialize the universal encoder $\mathcal{E}$ and decoder $\mathcal{D}$ using OFA-Large \cite{pengwang2022ofa}. The input image resolution is set to $512 \times 512$, with both input and output sequence lengths limited to $100$ tokens.  We use Adam as the optimizer with a batch size of $128$. The learning rate is initialized to $3 \times 10 ^{-5}$ with a linear decay strategy. Unlike existing works \cite{yuanzhengci2023unihcp}, our approach allows multiple tasks to coexist within the same batch, contributing to gradient stabilization. In single-task training, we conduct 4,000 iterations. In multi-task training, we ensure that the total number of training tokens for each task is approximately equivalent to that of single-task training.

\begin{table}[t]
\caption{Comparisons on \emph{RPar} task. We report mIOU metric on RefCIHP dataset.}
\centering
\begin{tabular}{l|c|c}
\hline
\textbf{Model} & {\textbf{Model Size}} & {\textbf{mIoU}} \\
\hline
Florence2-L \cite{binxiao2023florence} & 770M & 6.29 \\
Unified-IO2-L \cite{jiasen2024uio2} & 1.1B & 6.83 \\
\hline
\textbf{RefHCM} & 500M & \textbf{45.62} \\
\hline
\end{tabular}
\label{tab:rpar-result}
\end{table}

\subsection{Comparisons with State-of-the-arts}

In this section, we compare our RefHCM with state-of-the-arts on four referring human-centric tasks.

\noindent
\textbf{Main results on \emph{REC} task.} \noindent
Tab. \ref{tab:refcoco-results} summarizes the comparison results for \emph{REC} tasks on RefCOCO and RefCOCO+ testA datasets.  As shown, our model, RefHCM, demonstrates superior performance compared to models with similar parameter counts, such as UNITER \cite{yen2020uniter} and OFA-L-Multitask \cite{pengwang2022ofa}\footnote{OFA-L-tuned refers to OFA-L-Multitask model fine-tuned on RefCOCO series, effectively functioning as a single-task model.}. Even when compared to models with double the parameters, such as UNINEXT-H \cite{fangjian2023uninext},  RefHCM achieves results that are within a $1\%-2\%$ margin.  These results highlight the effectiveness of UniHCM in addressing the \emph{REC} task.

\emph{RKpt} task, which involves predicting both bounding boxes and keypoints, enables the utilization of bounding boxes derived from \emph{RKpt} task.
When limited  to the \emph{REC} task, which  predicts only bounding boxes, the performance drops by $4.03\%$  AP@50 on RefCOCO and $4.33\%$ AP@50 on RefCOCO+. This decline hightlights the significance  of incorporating human keypoint information  in enhancing human localization accuracy  (see Fig. \ref{fig:refpose-form}).

\vspace{0.4em}

\noindent
\textbf{Main results on \emph{RKpt} task.} \
To evaluate the \emph{RKpt} task, which traditional keypoint detection models typically do not support with language prompts, we adapt three multimodal language models for this evaluation: (1) \textbf{Unified-IO-2} \cite{jiasen2024uio2}: Unified-IO-2, a versatile  model that supports both \emph{REC} and Region Keypoints tasks, is adapted for \emph{RKpt}. The adaptation involve a two-step process: first, \emph{REC} task is  performed using the reference text to locate the target individual, and then the Region Keypoints task is applied to generate keypoints for the identified individual.
(2) \textbf{PoseGPT} \cite{yaofeng2023posegpt}: PoseGPT performs 2D keypoint detection by encoding keypoint coordinates as natural language. While the original PoseGPT leverages a significantly  larger foundation model (LLaVA), we ensure a fair comparison by training RefHCM using PoseGPT’s approach of representing keypoint coordinates as natural language,  denoted as $\text{PoseGPT}_\mathrm{text}$. (3) \textbf{UniPHD} \cite{uniphd2024miao}: This recent work introduces the RefHuman dataset and a pose-centric hierarchical decoder for  \emph{RKpt} task. 

The compared results on \emph{RKpt} task are shown in  Tab. \ref{tab:refpose-results} and Tab. \ref{tab:refhuman-results}. As shown in Tab. \ref{tab:refpose-results}, we can observe that: \textbf{(1)} RefHCM demonstrates robust performance, achieving over $70$ OKS AP across all three RefCOCO-series test sets. While Unified-IO-2 \cite{jiasen2024uio2}  attains only around $50$ OKS AP, highlighting the superiority of integrating  \emph{REC} and Region Keypoints task in RefHCM. \textbf{(2)} $PoseGPT_{\mathrm{text}}$ \cite{yaofeng2023posegpt} achieves comparable OKS performance to RefHCM but performs less effectively in terms of AP@50. 
 These results highlight the benefit of incorporating bounding box tokens as a preamble to keypoints, optimizing keypoint predictions
\textbf{(3)} Additionally, we evaluate RefHCM on the RefHuman dataset  \cite{uniphd2024miao} in a \textit{zero-shot} manner. As shown in Tab. \ref{tab:refhuman-results}, RefHCM achieves an OKS AP of $66.8$, marginally  surpassing  UniPHD's $66.7$. This result demonstrates the robust generalization capability of RefHCM in the \emph{RKpt} task.

 Fig. \ref{fig:rkpt-example} provides qualitative examples for the \emph{RKpt} task. As shown, RefHCM not only produces more precise predictions but also generates plausible  estimates for occluded keypoints.

\begin{table}[t]
\caption{Comparisions on  \emph{RHrc} task. We report CIDEr metric on CapCIHP dataset.}
\centering
\begin{tabular}{l|c|c}
\hline
\textbf{Model} & {\textbf{Model Size}} & {\textbf{CIDEr}} \\
\hline
Florence2-L \cite{binxiao2023florence} & 770M & 0.11 \\
Unified-IO2-L \cite{jiasen2024uio2} & 1.1B & 0.98 \\
LLaVA-v1.5-7b \cite{llava1.5} & 7B & 9.54 \\
\hline
\textbf{RefHCM} & 500M & \textbf{82.41} \\
\hline
\end{tabular}
\label{tab:rhrc-result}
\end{table}

\begin{figure}[t]
    \centering
    \includegraphics[width=0.48\textwidth]{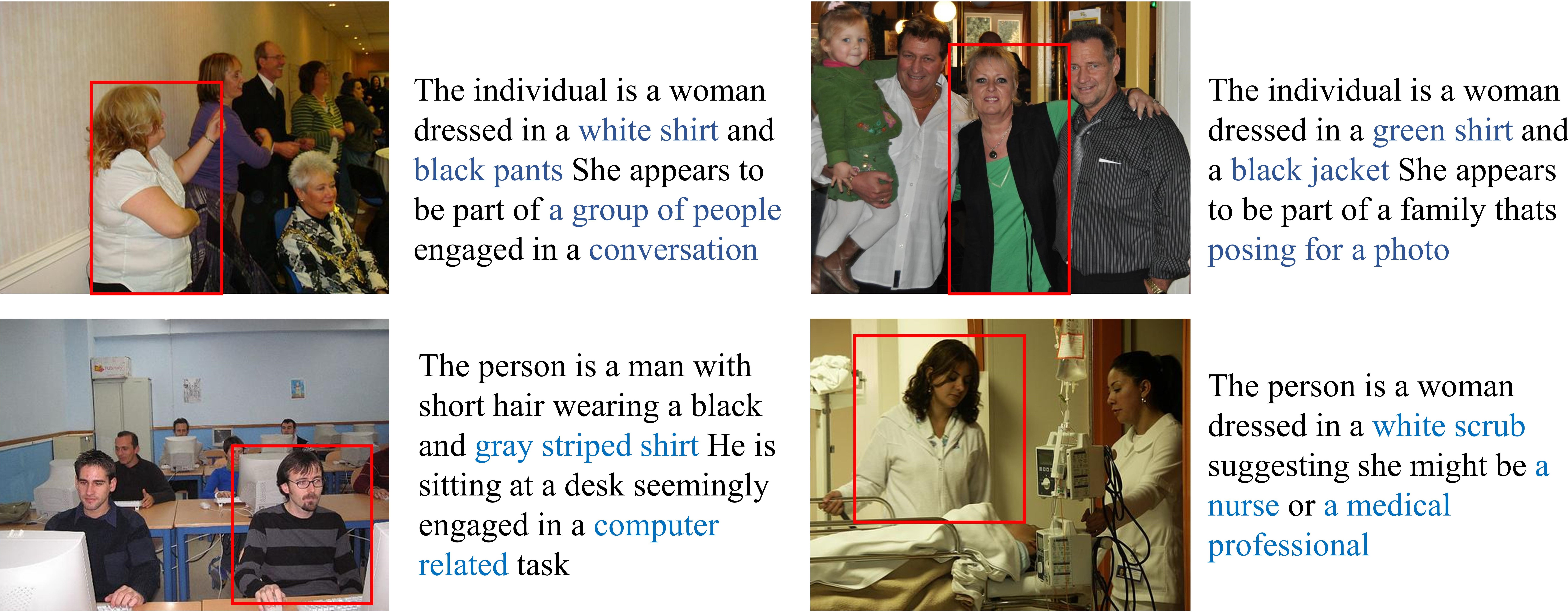}
    \caption{Qualitative results on \emph{RHrc} task. RefHCM generates more comprehensive captions that describe the target human's attire, position, identity, and other relevant attributes.}
    \label{fig:rcg-example}
\end{figure}

\vspace{0.4em}

\noindent
\textbf{Main results on \emph{RPar} task.} \
Unlike traditional parsing task, which typically requires a cropped image of the target individual, \emph{RPar} task  necessitates the use of the entire image along with language prompts to generate a parsing map of the target person. For \emph{RPar} evaluation, we adapt two multimodal language models: (1) \textbf{Florence2-L} \cite{binxiao2023florence}, which supports phrase segmentation, has has been used in ComfyUI\footnote{https://github.com/kijai/ComfyUI-Florence2} to generate human part masks for AI-generated content.  (2) \textbf{Unified-IO-2} \cite{jiasen2024uio2}, where we first perform \emph{REC} to obtain the target person's bounding box, and then use this box for region segmentation.  

The compared results for \emph{RPar} task are shown in  Tab. \ref{tab:rpar-result}.
As shown, RefHCM surpasses the other models by a large margin, showcasing   its effectiveness in referring human parsing. Qualitative examples are provided in Fig. \ref{fig:rpar-example}. Florence  \cite{binxiao2023florence} struggles in correctly identifying the target specified by the referring text, while Unified-IO-2 tends to segment the entire body instead of specific body parts. In constrast, RefHCM excels in accurately locating the target person specified by the referring text, and identifying corresponding body part regions.

\vspace{0.4em}

\noindent
\textbf{Main results on \emph{RHrc} task.} \
For \emph{RHrc} task, we compare RefHCM with multimodal language models: Florence2-L, Unified-IO2-L and LLaVA-v1.5-7b \cite{llava1.5}.
As shown in Tab \ref{tab:rhrc-result},  Florence2-L and Unified-IO2-L generate overly brief captions (e.g., "man" or "woman"), resulting in CIDEr scores below 1.0. While LLaVA-v1.5-7b can identify people and describe their clothing, it struggles with providing detailed perceptions of attires and activities. Fig \ref{fig:rcg-example} provides qualitative results, which demonstrates that RefHCM generates comprehensive, multi-aspect descriptions for the target individual.

\begin{table}[t]
\caption{Ablation study on single-task training versus multitask training of our RefHCM model.}
\centering
\begin{tabular}{l l l|c|l}
\hline
{\textbf{Task}} & {\textbf{Dataset}} & {\textbf{Metric}} & {\textbf{Single-Task}} & {\textbf{Multi-Task}} \\
\hline
\emph{REC} & Refcoco testA & AP@50 & 93.51 & \textbf{93.69} {\scriptsize(+0.18)} \\
\emph{REC} & Refcoco+ testA & AP@50 & \textbf{89.99} & 89.56 {\scriptsize(-0.43)} \\
\emph{RKpt} & Refpose testA & OKS AP & 72.78 & \textbf{75.60} {\scriptsize(+2.88)} \\
\emph{RKpt} & Refpose+ testA & OKS AP & 69.78 & \textbf{72.24} {\scriptsize(+2.46)} \\
\emph{RPar} & RefCIHP val & mIoU & 31.89 & \textbf{45.62} {\scriptsize(+\textbf{13.73})} \\
\emph{RHrc} &CapCIHP val & CIDEr & 79.09 & \textbf{82.41} {\scriptsize(+3.32)} \\
\hline
\end{tabular}
\label{tab:multitask-results}
\end{table}

\begin{table}[t]
\caption{Ablation study on different image tokenizer on \emph{RPar} task. We report mIoU for Reconstruction and Parsing. IR: Image Resolution; FT: Forward Times.}
\centering
\begin{tabular}{l|c|c|c|c}
\hline
{\textbf{Tokenizer}} & {\textbf{IR}} & {\textbf{FT}} & {\textbf{Reconstruction}} & {\textbf{Parsing}} \\
\hline
VQGAN \cite{patrick2021vqgan} & 256x256 & 256 & 72.20 & 17.01 \\
VQGAN-$S^2$ \cite{baifeng2024s2} & 128x96 & 48 & 73.20 & 35.08 \\
ViT-VQGAN \cite{jiahui22vitvqgan} & 256x256 & 256 & 62.16 & - \\
RQ-VAE \cite{doyup2022rqvae} & 256x256 & 1024 & 65.88 & - \\
TiTok \cite{qihang2024titok} & 128x96 & 32 & 61.73 & 18.7 \\
\hline
VQGAN-QPG & 128x96 & \textbf{1} & 73.20 & 37.50 \\
Modified VQGAN & 128x96 & 48 & \textbf{73.20} & \textbf{42.50} \\
\hline
\end{tabular}
\label{ablation:tokenizers}
\end{table}

\begin{table}[t]
\caption{Ablation study on different image compression strategies on \emph{RPar} task. We report the mIoU metric. PTN: Parsing Token number. }
\centering
\begin{tabular}{l|c| c|c}
\hline
{\textbf{Method}} & \textbf{PTN} & {\textbf{Reconstruction}} & {\textbf{Parsing}}\\
\hline
Whole Image & 256 & 72.18 & 17.01\\
Padding & 64 & \textbf{75.28} & 36.48\\
Resize & \textbf{48} &73.20 & \textbf{42.50}\\
\hline
\end{tabular}
\label{ablation:parsing}
\end{table}

\begin{table}[t]
\caption{Ablation study on different keypoint formulation strategys on \emph{RKpt} task.}
\centering
\begin{tabular}{l|rrr}
\hline
\multirow{2}{*}{\textbf{Method}} & \multicolumn{3}{c}{\textbf{Results(OKS AP)}} \\
 & Refpose & Refpose+ & Refposeg \\
\hline
Fixed kpts & 56.33 &    52.80 &    57.22 \\
Bbox + Fix kpts & 56.84 &54.04 & 57.00 \\
Named kpts & 62.88 & 60.97 & 64.79 \\
\hline
LCR & \textbf{72.78} & \textbf{69.78} & \textbf{75.69} \\
\hline
\end{tabular}
\label{ablation:keypoints}
\end{table}

\subsection{Ablation Study}

In this section, we conducted ablation studies to validate the effectiveness of our model RefHCM. The ablation results are shown in Tab. \ref{tab:multitask-results}-\ref{ablation:parsing}.

\noindent
\textbf{Effectiveness of multi-task training.} \
 Tab. \ref{tab:multitask-results} reports the performance of RefHCM when trained on single task versus multiple tasks. As shown, multi-task training generally outperforms single-task training, highlighting the advantages of unifying various human-centric referring tasks within a single model.
 However, multi-task training leads to a slight performance drop on the \emph{REC} task. This can be attributed to the varying distribution of bounding boxes across different tasks. For the \emph{REC} and \emph{RKpt} tasks, bounding box annotations are sourced from the COCO dataset, whereas \emph{RPar} and \emph{RHrc} tasks use bounding boxes extracted from individual masks. Additionally, we increase the sample rate for \emph{RPar} and conduct \emph{RHrc} pretraining in the multi-task training, further exacerbating the misalignment of bounding boxes.

\vspace{0.4em}

\noindent
\textbf{Ablation study on parsing merger/dispenser.} \
A modified VQGAN \cite{patrick2021vqgan} is used as the parsing merger/dispenser.  Tab. \ref{ablation:tokenizers} compares different image tokenizers, including original VQGAN \cite{patrick2021vqgan}, ViT-VQGAN \cite{jiahui22vitvqgan}, RQ-VAE \cite{doyup2022rqvae} and TiTok \cite{qihang2024titok}, for human parsing maps. As shown, our modified VQGAN achieves the best reconstruction and prediction performance. We also evaluate the performance of modified VQGAN combined with  QPG, which performs non-autogressive generation for parsing tokens. As shown in Tab. \ref{ablation:tokenizers},  VQGAN-QPG enables single-step generation of all parsing tokens, significantly reducing inference cost while retaining $88\%$ of original parsing performance. 

Notably, the predicted parsing maps only include individual person, necessitating their conversion  into full-image parsing maps. We compares  three strategies: \textit{Whole-Image Reconstruction}, which discretizes the parsing map within the whole image; \textit{Padding}, which pads the parsing maps to $128\times 128$ to match the the original VQGAN input size; \textit{Resize}, which directly resizes the generated parsing maps to match the size of predicted bounding box. As shown in Tab. \ref{ablation:parsing}, all three strategies achieve comparable reconstruction performance, while the \textit{Resize} strategy delivers the best parsing prediction performance.  This can be attributed to the \textit{Resize} strategy requiring the fewest parsing token predictions, thereby significantly reducing the complexity of parsing prediction and achieving superior results.

\vspace{0.4em}

\noindent
\textbf{Effectiveness of Location-Context Restriction (LCR).} \
Tab.  \ref{ablation:keypoints} reports the performance of RefHCM on \emph{RKpt} task. We compare differnt keypoint formulation strategies:  \textit{Fixed kpts}, which represents tokens of different keypoints in a fixed order; \textit{Bbox + Fix kpts}, which prepends the bounding box tokens to the token sequence of \textit{Fixed kpts}; \textit{Named kpts}, which inserts keypoint name before each keypoint tokens,
built on the  token sequence of \textit{Fixed kpts}; \textit{LCR}, which inserts keypoint names before each keypoint tokens, built on the  the token sequence of \textit{Bbox + Fix kpts}. As shown in Tab. \ref{ablation:keypoints}, the \textit{LCR} strategy achieves the best performance on OKS AP metric, with an increase of over $9\%$ compared to other keypoint formulation strategies. This significant improvement underscores the crucial role of semantic clues and the inclusion of bounding box information in keypoint prediction.

\begin{table*}[t]
\caption{Comparisons on \emph{ReasonDet} benchmark. AP@50 metric is reported.}
\centering
\begin{tabular}{l|c|c|c|c|c|c|c}
\hline
{\textbf{Model}} & {\textbf{Model Size}} & {\textbf{Total}} & {\textbf{Id}} & {\textbf{Pose/Clothing}} & {\textbf{Social}} & {\textbf{Physical}} & {\textbf{Future}} \\
\hline
MDETR \cite{ziyi2022mdetr} & 400M & 59.1 & 63.9 & 70.2 & 47.4 & 50.1 & 57.9 \\
OFA-L-Pretrain \cite{pengwang2022ofa} & 520M & 45.1 & 46.7 & 57.2 & 37.4 & 42.7 & 39.6 \\
QwenVL-chat-7B \cite{jinzebai2023qwenvl} & 7B & \textbf{69.8} & \textbf{71.9} & \textbf{76.3} & 59.1 & \textbf{69.9} & \textbf{68.7} \\
Ferret-13B \cite{haoxuanyou2024ferret} & 13B & 67.5 & 74.2 & 72.2 & \textbf{62.3} & 65.6 & 62.1 \\
\hline  
{RefHCM} & 500M & 55.9 & 58.6 & 65.7 & 45.4 & 55.0 & 52.8 \\
{RefHCM-tuned} & 500M & 65.5 & 64.4 & 70.8 & 61.1 & 66.0 & 64.3 \\
\hline
\end{tabular}
\label{tab:reasondec}
\end{table*}

\begin{table*}[t]
\caption{Comparisons on \emph{ReasonKpt} benchmark. OKS AP  metric is reported.}
\centering
\begin{tabular}{l|c|c|c|c|c|c|c}
\hline
{\textbf{Model}} & {\textbf{Model Size}} & {\textbf{Total}} & {\textbf{Id}} & {\textbf{Pose/Clothing}} & {\textbf{Social}} & {\textbf{Physical}} & {\textbf{Future}} \\
\hline
Unified-IO2-L \cite{jiasen2024uio2} & 1.1B & 50.2 & 57.4 & 55.8 & 42.7 & 42.7 & 53.3 \\
$PoseGPT_\text{text}$ \cite{yaofeng2023posegpt} & 500M & 58.7 & 59.7 & 66.6 & 44.6 & 49.0 & 56.3 \\
\hline
{RefHCM} & 500M & 57.6 & 57.8 & 63.6 & 43.6 & 52.9 & 53.2\\
{RefHCM-tuned} & 500M & \textbf{80.4} & \textbf{76.6} & \textbf{78.6} & \textbf{71.3} & \textbf{76.8} & \textbf{75.8} \\
\hline
\end{tabular}
\label{tab:reasonkpt}
\end{table*}

\begin{table*}[t]
\caption{Comparisons on \emph{ReasonPar} benchmark. mIoU metric is reported.}
\centering
\begin{tabular}{l|c|c|c|c|c|c|c}
\hline
{\textbf{Model}} & {\textbf{Model Size}} & {\textbf{Total}} & {\textbf{Id}} & {\textbf{Pose/Clothing}} & {\textbf{Social}} & {\textbf{Physical}} & {\textbf{Future}} \\
\hline
Unified-IO2-L \cite{jiasen2024uio2} & 1.1B & 5.1 & 5.1 & 5.0 & 5.3 & 4.8 & 5.2 \\
Florence2-L \cite{binxiao2023florence} & 770M & 5.8 & 6.2 & 5.8 & 5.5 & 5.8 & 5.8 \\
\hline
{RefHCM} & 500M & 22.8 & 22.4 & 26.5 & 20.0 & 23.2 & 21.7 \\
{RefHCM-tuned} & 500M & \textbf{26.6} & \textbf{26.6} & \textbf{27.6} & \textbf{24.1} & \textbf{28.6} & \textbf{25.3} \\
\hline
\end{tabular}
\label{tab:reasonpar}
\end{table*}

\subsection{Main Results on Reasoning Reference Benchmark}
In this section, we  compare the performance  on constructed Reasoning Reference Benchmark, \emph{ReasonRef}, which comprises three subset: \emph{ReasonDet}, \emph{ReasonKpt} and \emph{ReasonPar}.

\noindent
\textbf{Main results on \emph{ReasonDet} benchmark.} \
We compare RefHCM with two open-source multimodal language models, Ferret-13B \cite{haoxuanyou2024ferret} and QWen-VL \cite{jinzebai2023qwenvl}, and two similarly sized models, MDETR \cite{ziyi2022mdetr} and OFA \cite{pengwang2022ofa}, on \emph{ReasonDet} benchmark for \emph{REC} task.  As shown in Tab. \ref{tab:reasondec}, with a comparable model size, RefHCM significantly outperforms  MDETR \cite{ziyi2022mdetr} and OFA \cite{pengwang2022ofa}.  Meanwhile, Ferret-13B \cite{haoxuanyou2024ferret} and QWen-VL \cite{jinzebai2023qwenvl}, powered by larger LLMs, demonstrate a clear advantage in reasoning capabilities.
 All models perform relatively well in the Identity and Pose/Clothing dimensions but faced challenges in the Social Relation, Physical Relation, and Future Prediction dimensions, which require   more sophisticated reasoning abilities. 
Encouragingly, after fine-tuning on \emph{$\text{ReasonRef}_{\text{train}}$} (denoted as RefHCM-tuned), the total score improves from $55.9$ to $65.5$, approaching  the performance level of Ferret-13B. A detailed analysis of subcategories reveals that the dimensions of Social Relation, Physical Relation, and Future Prediction exhibit the most substantial improvements. This highlights the potential of targeted fine-tuning in narrowing the performance gap between our smaller RefHCM model and larger multimodal language models.

\vspace{0.4em}

\noindent
\textbf{Main Results on \emph{ReasonKpt} and  \emph{ReasonPar} benchmark.} \
Tab. \ref{tab:reasonkpt} compares RefHCM with Unified-IO2-L \cite{jiasen2024uio2} and $PoseGPT_\text{text}$ \cite{yaofeng2023posegpt}  on \emph{ReasonKpt} benchmark for \emph{RKpt} task. 
As shown, RefHCM outperforms Unified-IO2-L and achieves comparable performance to $PoseGPT_\text{text}$. After fine-tuning on \emph{$\text{ReasonRef}_{\text{train}}$}, RefHCM significantly  surpasses  both Unified-IO2-L and $PoseGPT_\text{text}$, particularly excelling in the dimensions of social relation and future predictions.
Tab. \ref{tab:reasonkpt} also compares RefHCM with Unified-IO2-L \cite{jiasen2024uio2} and Florence2-L \cite{binxiao2023florence}  on \emph{ReasonPar} benchmark for \emph{RPar} task. 
As shown, RefHCM outperforms both Unified-IO2-L and Florence2-L by a large margin, demonstrating its superiority in reasoning for referring parsing. 

%% file: 5_conclusion.tex
\section{Conclusion}

In this paper, we introduce referring human perceptions, a novel task aimed at predicting human perceptions including locations, poses, and parsing for specified individuals based on natural, user-friendly text. 
This innovation not only holds significant potential for advancing human-AI interactions in areas like chatbots and sports analysis, but also enhances the utility and effectiveness of AI-generated content by providing prior information like captions, keypoints and masks of the target person for generation models. By combining robust referring capabilities with advanced perception understanding, existing AI systems can markedly improve their ability to follow complex instructions involving humans, leading to more effective task execution.
Our proposed RefHCM, a unified model for referring human perception, achieves top-tier performance on this task and on the new \emph{ReasonRef} benchmark, setting a solid foundation for future research. We believe that our proposed task, model, and benchmark will inspire further advancements in human-AI interaction, paving the way for effective AI systems.